\documentclass[11pt]{IEEEtran}

\usepackage{cite}
\usepackage{amsmath,amssymb,amsfonts}
\usepackage{algorithmic}
\usepackage{graphicx}
\usepackage{textcomp}
\usepackage{xcolor}
\usepackage{url}
\usepackage{geometry}
\usepackage{enumitem}
\usepackage{tikz}
\usepackage{threeparttable}
\def\BibTeX{{\rm B\kern-.05em{\sc i\kern-.025em b}\kern-.08em
    T\kern-.1667em\lower.7ex\hbox{E}\kern-.125emX}}

\makeatother
\makeatletter

\newcommand\copyrighttext{%
  \footnotesize \textcopyright 2026 IEEE. Personal use of this material is permitted.
  Permission from IEEE must be obtained for all other uses, in any current or future
  media, including reprinting/republishing this material for advertising or promotional
  purposes, creating new collective works, for resale or redistribution to servers or
  lists, or reuse of any copyrighted component of this work in other works.}
\newcommand\copyrightnotice{%
\begin{tikzpicture}[remember picture,overlay]
\node[anchor=south,yshift=10pt] at (current page.south) 
  {\fbox{\parbox{\dimexpr\textwidth-\fboxsep-\fboxrule\relax}{\copyrighttext}}};
\end{tikzpicture}%
}
\begin{document}
\setlength{\columnsep}{0.24in}
\title{Large Language Model Deployment on Resource-Constrained Edge Devices: A Practitioner's Survey}
\author{
  \begin{minipage}[t]{0.3\textwidth}
    \centering
    \IEEEauthorblockN{Rahul Maliakkal}\\
    \IEEEauthorblockA{rahuljm@meta.com}
  \end{minipage}
  \hfill
  \begin{minipage}[t]{0.3\textwidth}
    \centering
    \IEEEauthorblockN{Yashasvi Makin}\\
    \IEEEauthorblockA{yashasvi@meta.com}
  \end{minipage}
  \hfill
  \begin{minipage}[t]{0.3\textwidth}
    \centering
    \IEEEauthorblockN{Plawan Rath}\\
    \IEEEauthorblockA{plawan@meta.com}
  \end{minipage}
  \vspace{1em} 
  \begin{minipage}[t]{0.33\textwidth}
    \centering
    \IEEEauthorblockN{Rohit Jain}\\
    \IEEEauthorblockA{rkjain@meta.com}
  \end{minipage}
  \hspace{0.2em} 
  \begin{minipage}[t]{0.33\textwidth}
    \centering
    \IEEEauthorblockN{Ankur Sadhoo}\\
    \IEEEauthorblockA{asadhoo@meta.com}
  \end{minipage}
}
\maketitle
\copyrightnotice
\IEEEpubidadjcol

\begin{abstract}
Large Language Models (LLMs) have transformed many areas of artificial intelligence (AI), but their substantial resource requirements limit their deployment on mobile and edge devices. Although LLMs have demonstrated impressive capabilities in various applications, the next challenge is deploying these models on resource-limited edge devices. Recent innovations in software optimizations and specialized hardware have made it feasible to deploy LLMs on edge devices. Edge deployment offers benefits including local data processing, cost savings, reduced latency, and personalized responses. This paper reviews the latest techniques that make the deployment of LLMs on edge devices feasible.
\end{abstract}

\begin{IEEEkeywords}
Large Language Models, Edge Computing, Model Fine-tuning, Parameter-Efficient Fine-Tuning, Model Compression, Quantization, Pruning, Knowledge Distillation, Inference Optimization, Edge AI, On-Device AI, Model Deployment, Resource-Constrained Devices, Low-Latency Inference
\end{IEEEkeywords}

\section{Introduction}
Large Language Models (LLMs)\cite{b1} are ubiquitous these days and power everything from conversational interfaces like ChatGPT \cite{b2} to content generation \cite{b3,b4} These models have traditionally relied on cloud-based inference, where data is sent from the edge device and processed on central cloud servers. This approach offers significant advantages in terms of scalability and access to powerful compute resources. However, it also introduces potential challenges, such as issues with data residency \cite{b5}, server costs, and reliance on stable network connectivity and low latency \cite{b6}.

While individual research has explored model compression \cite{b45}, fine-tuning\cite{b59}, and inference optimization, our work provides a comprehensive perspective by unifying these techniques and explicitly incorporating hardware and runtime considerations. This integrated approach is crucial for practical deployment of LLMs on resource-constrained edge devices. By synthesizing software optimizations with hardware-aware strategies, we enable practitioners to balance model accuracy, latency and memory usage. Furthermore, we offer actionable deployment guidelines that address real-world constraints, ensuring that LLMs can be effectively adapted and optimized for diverse edge environments.

Edge devices have unique characteristics that include memory constraints, heterogeneous hardware platforms, and strict energy budgets. In Section II, we discuss the key challenges that require specialized methods for deploying LLMs on the edge. Then, in Section III, we outline different techniques for efficient fine-tuning that conserve resources. Section IV focuses on software strategies, including model compression and optimization methods for inference. This section also focuses on hardware and runtime considerations for LLM deployment on edge devices. In Section V, we provide practical deployment guidelines synthesizing techniques for practitioners. Section VI reviews future research directions and open challenges in the field. Section VII presents a conclusion of this survey.

\section{Challenges in Deploying LLMs at the Edge}
Most edge devices typically have limited computational resources, and stringent energy requirements, creating significant deployment challenges. 

\subsection{Memory Constraints}
Edge devices often operate with very limited memory capacity, typically ranging from 1-4 GB of RAM \cite{b54} in most smartphones and edge devices. In contrast, modern LLMs have substantial memory requirements that exceed these limitations. For example, the Mistral-7B model, which is considered relatively small, still needs 13 GB of memory for full-precision inference. This requirement far exceeds the available memory on nearly all consumer edge devices.

\subsection{Heterogeneous Hardware}
Edge computing is characterized by a large variety of hardware architectures, with examples ranging from ARM CPUs like the Cortex-A72 (Raspberry Pi 4), mobile GPUs (Adreno 730), and specialized AI accelerators like Edge TPUs \cite{b55}. This diversity and heterogeneity require models to be adapted to diverse instruction sets, memory hierarchies, and compute capabilities. For example, Mistral-7B's performance shows significant variation in energy efficiency between different hardware architectures \cite{b56}.

\subsection{Model Size}
Another constraint of edge hardware platforms is storage size, with typical devices having less than 32 GB of total storage with only a fraction available for AI workloads \cite{b54}, leading to challenges due to the substantial storage capacity demands of LLMs. For example, Mistral-7B's raw model weights occupy 13 GB in FP16 and 7 GB in 8-bit quantization \cite{b52}.

\subsection{Energy Consumption}
Inference with LLMs can be energy-intensive, which impacts device longevity and operational costs. For example, running Mistral-7B on a Raspberry Pi 4 (4GB RAM, 5W TDP) takes over 30 seconds and consumes up to 0.04 Wh per inference \cite{b57}. On mobile devices, running even quantized LLMs can drain 2–5\% of the battery per minute of continuous inference \cite{b58}. 

\section{Resource Efficient Fine-tuning}

\subsection{Edge Training Limitations and the Pre-trained Model Paradigm}
Training LLMs on-device is impractical due to memory requirements and heterogeneous hardware constraints \cite{b8,b9}. We therefore focus on transfer learning with pre-trained models from Hugging Face and TensorFlow Hub, optimized via techniques discussed in the following sections

\subsection{Model Fine-tuning}
Model fine-tuning involves adapting a pre-trained model using a smaller, context-specific
dataset \cite{b59} to enhance its performance on a particular task or domain. Fine-tuning enables models to be updated and customized directly on the device, which is essential in offline systems that lack cloud connectivity. This enables the model to give accurate and contextually aware responses even when the device is not connected to the internet.

On-device fine-tuning offers significant advantages. Communication with remote servers is no longer needed, which reduces response time. Real-time applications become more responsive as a result. The model can adapt and learn based on local usage patterns over time. Adapting LLMs for edge deployment presents several challenges due to resource limitations. Edge devices usually possess restricted computational power and limited memory capacity. This makes traditional full fine-tuning methods difficult to implement on these devices. Despite these limitations, modern fine-tuning strategies, such as Parameter-Efficient Fine-Tuning (PEFT) techniques, have been developed to address these issues\cite{b11,b60,b61}. PEFT considerably reduces the computational and memory footprint associated with adapting LLMs to the edge. In contrast to full fine-tuning, which requires extensive resources and updates all model parameters, PEFT methods adjust only a small subset of parameters or add lightweight adapters, enabling efficient customization with minimal overhead. (usually 0.1-3\% of total parameters) while maintaining performance comparable to full fine-tuning.

LoRA \cite{b12} is the most widely used PEFT method. It divides weight updates into two smaller low-rank matrices that are trained without changing the initial model weights. This enables quick task adaptation and preserves model performance while reducing trainable parameters by 99\%. QLoRA \cite{b13} is an advanced variant that further reduces memory requirements by 75\% by combining LoRA with 4-bit quantization. One sophisticated PEFT technique for LLMs is DoRA (Weight-Decomposed Low-Rank Adaptation)\cite{b14}. By separating the pretrained weight matrix into magnitude and direction, it improves upon the original LoRA method. DoRA then focuses on the adaptation of the direction component, which improves performance and stability during fine-tuning

\section{Resource Efficient Inference}
\subsection{Model Compression and Optimization}
Model compression techniques are essential for deploying LLMs in edge environments, where computational resources and memory are limited. Edge devices are typically constrained by limited memory, often less than 8 GB, which imposes a tight budget for deploying LLMs. To enable efficient on-device inference various model compression techniques have been introduced to significantly lower the memory footprint. The most commonly used model compression methods are quantization, pruning, and distillation

\subsubsection{Quantization}
Quantization is a model compression technique that reduces the precision of model weights and activations from 16-bit or 32-bit floating-point representations to lower-bit formats such as 8-bit, 4-bit, or even 2-bit integers \cite{b18}. This precision reduction directly translates to smaller model sizes and faster inference speeds. However, the model's performance can decline due to loss of numerical precision. When model parameters are represented with fewer bits, some information is inevitably lost, potentially impacting model performance. Various techniques exist to mitigate this precision loss. In this section, we examine various quantization strategies specifically designed for edge deployments.

Post-Training Quantization (PTQ) converts full-precision neural networks to lower-precision formats such as 8-bit or 4-bit without requiring additional training resources. This makes it highly suitable for deployment in resource-constrained edge environments. GPTQ \cite{b15} applies inverse Hessian information for layer-by-layer quantization. The method compresses models to 4-bit precision with minimal performance loss by prioritizing weights based on their significance. This approach minimizes quantization errors by considering the curvature properties of the loss function during compression. SmoothQuant \cite{b16}, developed
at MIT, migrates quantization difficulty from activations to weights, achieving a 1.56x speedup and 2x memory reduction. Additionally, Activation-aware Weight Quantization (AWQ) \cite{b17} introduces activation-aware techniques to further reduce quantization error, enhancing performance in LLMs.

Another major method for neural network quantization is Quantization-Aware Training (QAT). Unlike PTQ, QAT incorporates quantization effects directly into the training process, allowing the model to adapt to lower-precision representations such as 8-bit or 4-bit during optimization. QAT enables the network to learn parameters that are robust to quantization-induced errors by simulating quantization during both the forward and backward passes. This often results in higher accuracy compared to PTQ, especially for very low bit-widths \cite{b18}. QAT is particularly beneficial when deployment scenarios demand minimal accuracy loss and the resources for retraining are available. It has been widely adopted in both computer vision and natural language processing tasks, and is supported by major deep learning frameworks. Recent advances in QAT have further improved its efficiency and scalability, making it a preferred choice for production-level quantized models.

\subsubsection{Pruning}
Pruning is a commonly used technique for model compression that aims to reduce the
complexity of a trained model by removing parameters that have minimal impact on its
performance. These parameters can consist of individual weights, entire neurons, or even
complete layers. The goal of pruning is to develop a sparser model that retains accuracy similar to that of the original full-sized network. This approach enables faster inference times and reduces memory usage \cite{b7}.

Pruning methods are generally categorized into unstructured and structured approaches. Structured pruning, eliminates entire neurons, channels, or layers, producing models that are more compatible with standard hardware and easier to accelerate \cite{b23}. This approach maintains the original model structure while reducing computational complexity. This makes it particularly suitable for deployment on edge devices where inference speed is critical. Unstructured pruning, in contrast, eliminates individual weights according to their importance or size, producing sparse weight matrices that enable significant compression. However, efficiently executing these sparse models typically depends on specialized hardware \cite{b19,b22}. 

Recent advances have introduced highly effective pruning techniques for LLMs. For example, SparseGPT \cite{b20} applies unstructured pruning to transformer models by leveraging second-order information, as a result of this we get significant sparsity (up to 90\%) with minimal accuracy loss. WAnda \cite{b21} uses activation-aware pruning strategies to further improve the efficiency and performance of pruned models, particularly in resource-constrained deployment scenarios.

Pruning can be performed in two ways. Iterative pruning during training gradually removes
parameters and retrains the model to recover accuracy. One-shot pruning applies a pruning mask in a single step, often followed by fine-tuning. Combining pruning with other compression techniques like quantization can lead to substantial reductions in model size and inference latency with minimal impact on accuracy \cite{b7}.

\subsubsection{Knowledge distillation}

Knowledge distillation is a model compression technique in which a smaller, more efficient
``student'' model is trained to replicate the behavior of a larger, high-performing ``teacher'' model. The student model learns from the ground-truth labels as well as from the soft predictions such as logits or probability distributions generated by the teacher. These predictions contain more detailed information about the teacher's learned representations and decision boundaries  \cite{b24}.

This method allows the student model to achieve competitive accuracy while using fewer parameters and less computing power. This smaller resource requirement makes it possible to deploy on edge devices with limited processing, storage, and battery life. Knowledge distillation has been effectively applied in the compression of LLMs for edge and mobile device applications \cite{b25,b26}.

There are several variants of knowledge distillation, such as response-based distillation, where the student mimics the teacher's output probabilities, and feature-based distillation, where intermediate representations are also transferred. Recent research has explored additional ways to improve the efficiency and generalization of student models. They include the distillation for transformer architectures, multi-teacher setups, and task-specific adaptations \cite{b27}. Knowledge distillation is frequently combined with other compression methods like quantization and pruning. It enhances model efficiency while preserving high accuracy.

\subsection{Efficient Decoding Techniques}
Edge applications require aggressive inference time optimization due to strict compute, battery and latency constraints\cite{b28}. LLM inference optimizations for edge deployments include memory-efficient techniques and parallelization strategies\cite{b29}. 

Key-Value(KV) caching stores previously computed attention matrices to eliminate recomputation during auto-regressive generation. This reduces complexity from O(n²) to O(n) and can achieve a speedup of up to 5×. Further optimizations, such as Paged attention, dynamically allocate variable-length sequences in non-contiguous memory blocks for optimal memory utilization on resource-constrained devices  \cite{b30,b31}.

Parallelization strategies enable the simultaneous processing of multiple requests, thereby making model serving faster. Techniques such as speculative decoding use smaller draft models to generate multiple candidate tokens, which are then verified in parallel by the target model. This approach maintains output quality while achieving 2-4x faster inference. More advanced methods, like adaptive draft length, further improve efficiency by dynamically adjusting the amount of speculation based on how often verifications succeed \cite{b32,b33}.

\subsection{Hardware and Runtime Considerations}
Deploying LLMs on edge devices requires careful attention to hardware limitations and the
selection of efficient runtime engines \cite{b34,b35}. The heterogeneous landscape of edge hardware across multiple manufacturers and architectures demands specialized model serving solutions capable of managing fluctuating workloads while maintaining performance\cite{b35,b36}. 

Local frameworks like Ollama, LM Studio, and GPT4All make it easier to install and set up
software on consumer hardware \cite{b37,b38,b39}.

High-performance inference engines like vLLM and SGLang employ memory optimizations like Paged Attention and Radix Attention to maximize throughput and memory efficiency. These inference engines also use parallelization strategies like dynamic batching and speculative decoding to serve multiple requests simultaneously \cite{b40,b41}. This makes model serving more faster and cost effective. 

Cross-platform runtimes like ONNX Runtime and ExecuTorch support diverse hardware
accelerators and quantized models, thereby enabling efficient inference on CPUs, GPUs, and specialized AI chips \cite{b42,b43}. For large-scale deployments, Kubernetes is increasingly used for orchestrating containerized model serving at the edge, providing automated scaling and fault tolerance \cite{b44}.

Hardware considerations are central to edge deployment. Memory bandwidth (often 2-4× model size), compute unit utilization, and power constraints drive precision trade-offs between FP32, FP16, INT8, and FP4 formats \cite{b35,b46}. Efficient model formats like GGUF and quantized ONNX help reduce the model's memory footprint, which makes it possible to run on devices with limited resources \cite{b47,b48}. Edge devices are increasingly utilizing specialized accelerators like TPUs and NPUs in addition to CPUs and GPUs to speed up inference and use less energy. Tensor Processing Units (TPUs) are specifically designed for deep learning tasks. They can cut down on latency and power significantly, making them perfect for real-time LLM inference on mobile and embedded platforms \cite{b49}. Neural Processing Units (NPUs) are special chips built into devices like Qualcomm Snapdragon processors or designed by ARM. They help run AI tasks, like neural networks, faster and use less power right on the device \cite{b50,b51}. Platforms like NVIDIA Jetson, which feature integrated AI-optimized GPUs with tensor cores, deliver comparable performance and efficiency, significantly enhancing edge AI capabilities. Support for these accelerators is growing in modern runtimes, allowing models to take advantage of hardware-specific optimizations for improved performance and efficiency. Recent work on hardware-software co-design has further demonstrated the importance of considering both hardware architectures and software optimizations together for sustainable and efficient edge deployments\cite{b53}.

\begin{figure}[!ht]
  \centering
  \includegraphics[width=3in]{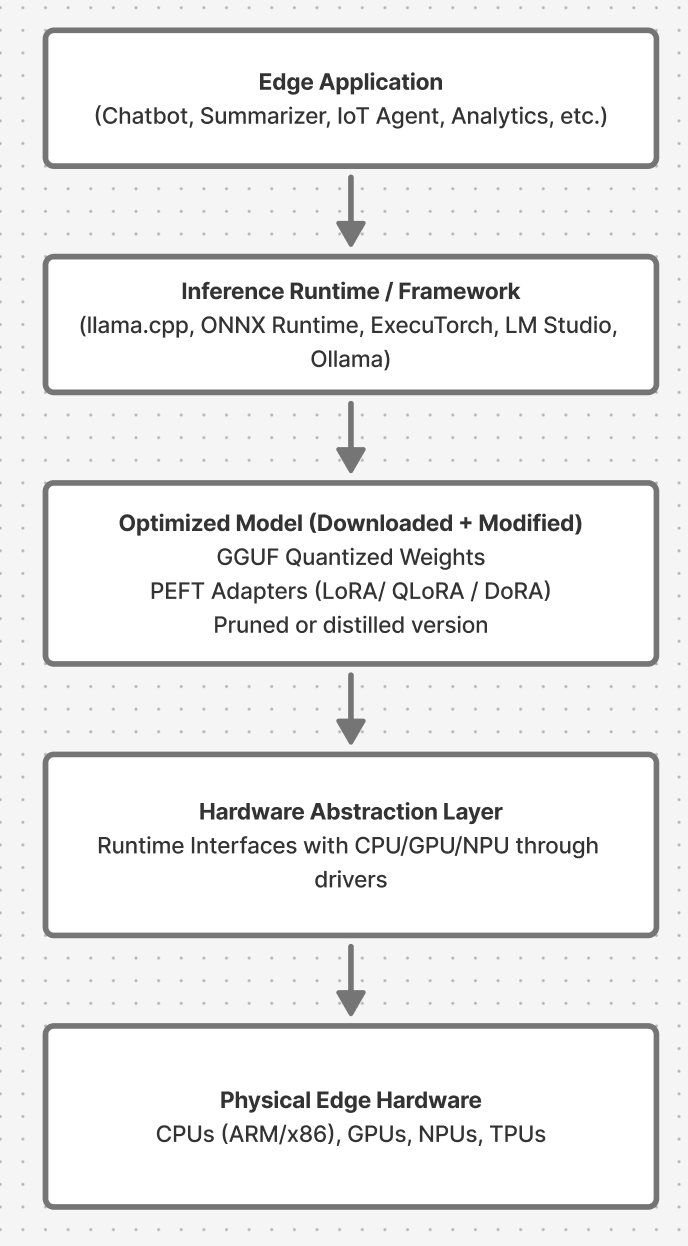}
  \caption{Edge LLMs combine efficient inference and fine-tuning. These are supported by specialized runtimes and hardware adaptations.}
  \label{fig:overview}
\end{figure}

\section{Practical Deployment Guidelines for Edge LLMs}
Deploying LLMs on edge devices requires navigating a range of trade-offs between model size, latency, accuracy, and device constraints. Table 1 provides a comprehensive overview of recommended optimization strategies across different device classes, categorized by available memory. Based on the surveyed techniques and current literature, the following guidelines are offered for practitioners.
\subsection{Model Size and Memory Budget}
Memory is the dominant constraint for edge deployment. A 7B model in FP16 requires
approximately 13\,GB RAM, while INT8 quantization reduces this to approximately
7\,GB. Only with 4-bit quantization does the memory footprint drop to around 4\,GB, making it feasible for smartphones with 4\,GB RAM. Sub-7B models or aggressively compressed 7B models are recommended for devices with limited memory.
\subsection{Compression and Accuracy Trade-offs}
Quantization techniques such as GPTQ and QLoRA offer 2--4$\times$ memory and
compute savings with minimal accuracy degradation. Structured pruning is hardware-friendly and improves inference speed, while unstructured pruning, such as SparseGPT, yields higher compression rates at the cost of runtime
complexity~\cite{b20}. Knowledge distillation (e.g., DistilBERT, TinyBERT) effectively compresses models while retaining accuracy, making them suitable for mobile deployment~\cite{b25,b26}.
\subsection{Fine-tuning and Adaptation}
Full fine-tuning is impractical on most edge devices due to high memory and compute demands. PEFT techniques such as LoRA and QLoRA reduce trainable parameters to less than 1--3\% of the full model~\cite{b12,b13}. These techniques enable on-device personalization or task-specific adaptation without requiring full retraining.
\subsection{Runtime and Platform Considerations}
Quantization-aware runtimes are essential. Frameworks such as ONNX
Runtime and \texttt{llama.cpp} provide robust support for INT4/INT8 models across diverse hardware accelerators including CPUs and NPUs. Format choice matters: GGUF and ONNX reduce load time and memory overhead, improving inference performance on resource-constrained systems.
\subsection{Energy and Latency Optimization}
Inference with LLMs is energy-intensive. Techniques such as key-value (KV) caching and paged attention help reduce inference latency and memory pressure by reusing computation. Emerging methods like speculative decoding further
improve responsiveness under tight compute budgets. Practitioners should evaluate
runtime behavior in low-power states and consider applying mixed-precision or dynamic
quantization based on thermal or battery constraints
\renewcommand{\arraystretch}{1.5}
    \begin{table}[t]
\centering
\caption{Edge Device Class Optimization Guidelines}
\label{tab:edge_device_optimization}
\begin{tabular}{|l|c|c|c|c|}
\hline
\textbf{Device Class} & \textbf{$<$ 2GB} & \textbf{2-4GB} & \textbf{4-8GB} & \textbf{$>$ 8GB} \\
\hline
\hline
Device & ESP32, & Budget & Mid-range & Jetson, \\
Examples & RPi Zero & phones & phones & High-end \\
\hline
Model & TinyLlama & Phi-2 & Mistral & Llama \\
Examples & 1.1B & 2.7B & 7B & 13B \\
\hline
Quantization & INT4/2 & INT4 & INT4/8 & INT8/16 \\
\hline
Pruning & 70-90\% & 50-70\% & 20-50\% & 0-20\% \\
\hline
Memory & 0.5-1.5 & 1.5-3.5 & 3.5-7 & 7-15 \\
(GB) & & & & \\
\hline
KV Cache & 256-512 & 512-1K & 2-4K & 4K+ \\
(tokens) & & & & \\
\hline
Energy & Low & Medium & Medium & High \\
(Watts) & 0.5-2 & 2-5 & 5-15 & 15-30 \\
\hline
\end{tabular}
\end{table}

\section{Future Work}
\begin{enumerate}
    \item \textbf{Adaptive Model Compression Techniques:} Current model compression techniques use uniform strategies throughout entire models. However, edge devices have varying resource availability that depends on factors such as battery levels and thermal conditions. Future research should focus on creating adaptive compression methods that can adjust quantization levels and pruning ratios in real-time based on the specific constraints of the device.
    \item \textbf{Cross-Platform Hardware Optimization:} As edge hardware diversity grows, creating standardized, hardware-agnostic runtimes and compilers that can automatically optimize LLMs for heterogeneous architectures (CPUs, GPUs, NPUs, TPUs, FPGAs) will improve deployment scalability and performance.
    \item \textbf{Neural Architecture Search for Edge LLMs:} Develop automated Neural Architecture Search (NAS) frameworks with hardware-aware search spaces to discover optimal edge-native LLM architectures. By incorporating multi-objective optimization across memory, latency, and accuracy constraints, NAS can automatically identify architectures that surpass manually designed compressed models, building on recent platform-aware approaches for resource-constrained devices \cite{b62}.
    \item \textbf{Continuous On-Device Learning with Minimal Computational Overhead:} Design ultra-lightweight learning algorithms that enable LLMs to continuously adapt to user behavior and local data patterns in real-time. This would allow models to provide increasingly personalized responses while operating within the strict computational and memory constraints of edge devices.
\end{enumerate}

\section{Conclusion}

This paper provides a thorough review of techniques for deploying pre-trained LLMs on resource-constrained edge devices. The main challenges associated with deploying LLMs in these environments that we examined include memory limitations, heterogeneous hardware, model size, and energy consumption. As shown in Fig.~\ref{fig:overview}, to effectively deploy LLMs at the edge, we must consider a unified approach to both the software and hardware aspects of deployment. We emphasize fine-tuning methodologies, model compression techniques, and efficient decoding algorithms in the software considerations. The hardware considerations concentrate on optimizations specific to the platform and runtime environments. By integrating these solutions, we can significantly reduce model size and memory requirements while maintaining competitive accuracy. It enables practical deployment on resource-constrained edge devices. As advancements in dedicated edge hardware and software optimization continue, we anticipate a substantial shift towards the widespread deployment of intelligent systems that utilize LLM functionalities directly on edge devices.


\begin{thebibliography}{00}
\bibitem{b1} Brown, T., Mann, B., Ryder, N., Subbiah, M., Kaplan, J.D., Dhariwal, P., Neelakantan, A., Shyam, P., Sastry, G., Askell, A. and Agarwal, S., 2020. Language models are few-shot learners. Advances in neural information processing systems, 33, pp.1877-1901.
\bibitem{b2} Achiam, J., Adler, S., Agarwal, S., Ahmad, L., Akkaya, I., Aleman, F.L., Almeida, D., Altenschmidt, J., Altman, S., Anadkat, S. and Avila, R., 2023. Gpt-4 technical report. arXiv preprint arXiv:2303.08774.
\bibitem{b3} Team, G., Anil, R., Borgeaud, S., Wu, Y., Alayrac, J.B., Yu, J., Soricut, R., Schalkwyk, J., Dai, A.M. and Hauth, A., 2024. Gemini: A family of highly capable multimodal models, 2024. arXiv preprint arXiv:2312.11805, 10.
\bibitem{b4} Hong, W., Ding, M., Zheng, W., Liu, X. and Tang, J., 2022. Cogvideo: Large-scale pretraining for text-to-video generation via transformers. arXiv preprint arXiv:2205.15868.
\bibitem{b5} Yao, Y., Duan, J., Xu, K., Cai, Y., Sun, Z. and Zhang, Y., 2024. A survey on large language model (llm) security and privacy: The good, the bad, and the ugly. High-Confidence Computing, 4(2), p.100211.
\bibitem{b6} Qu, G., Chen, Q., Wei, W., Lin, Z., Chen, X. and Huang, K., 2025. Mobile edge intelligence for large language models: A contemporary survey. IEEE Communications Surveys and Tutorials.
\bibitem{b7} Janusz, M., Wojnar, T., Li, Y., Benini, L. and Adamczewski, K., 2025. One Shot vs. Iterative: Rethinking Pruning Strategies for Model Compression. arXiv e-prints, pp.arXiv-2508.
\bibitem{b8}J. Devlin, M.-W. Chang, K. Lee, and K. Toutanova, "BERT: Pre-training of deep bidirectional transformers for language understanding," arXiv preprint arXiv:1810.04805, 2018.
\bibitem{b9}M. Sandler, A. Howard, M. Zhu, A. Zhmoginov, and L.-C. Chen, "MobileNetV2: Inverted Residuals and Linear Bottlenecks," CVPR, 2018.
\bibitem{b10}Han, X., Zhang, Z., Ding, N., Gu, Y., Liu, X., Huo, Y., Qiu, J., Yao, Y., Zhang, A., Zhang, L. and Han, W., 2021. Pre-trained models: Past, present and future. Ai Open, 2, pp.225-250.
\bibitem{b11} Hu, Z., Wang, L., Lan, Y., Xu, W., Lim, E.P., Bing, L., Xu, X., Poria, S. and Lee, R.K.W., 2023. Llm-adapters: An adapter family for parameter-efficient fine-tuning of large language models. arXiv preprint arXiv:2304.01933.
\bibitem{b12} Hu, E.J., Shen, Y., Wallis, P., Allen-Zhu, Z., Li, Y., Wang, S., Wang, L. and Chen, W., 2022. Lora: Low-rank adaptation of large language models. ICLR, 1(2), p.3.
\bibitem{b13} Dettmers, T., Pagnoni, A., Holtzman, A. and Zettlemoyer, L., 2023. Qlora: Efficient finetuning of quantized llms. Advances in neural information processing systems, 36, pp.10088-10115.
\bibitem{b14} Liu, S.Y., Wang, C.Y., Yin, H., Molchanov, P., Wang, Y.C.F., Cheng, K.T. and Chen, M.H., 2024, July. Dora: Weight-decomposed low-rank adaptation. In Forty-first International Conference on Machine Learning.
\bibitem{b15} Frantar, E., et al. "GPTQ: Accurate Post-Training Quantization for Generative Pre-trained Transformers." NeurIPS 2022.
\bibitem{b16} Xiao, T., et al. "SmoothQuant: Accurate and Efficient Post-Training Quantization for Large Language Models." ICML 2023.
\bibitem{b17} Lin, S., et al. "AWQ: Activation-aware Weight Quantization for LLM Compression and Acceleration." arXiv preprint arXiv:2306.00978, 2023.
\bibitem{b18} Jacob, B., et al. "Quantization and Training of Neural Networks for Efficient Integer-Arithmetic-Only Inference." CVPR 2018.
\bibitem{b19} Han, S., et al. "Learning both Weights and Connections for Efficient Neural Networks." NIPS 2015.
\bibitem{b20} Frantar, E., et al. "SparseGPT: Massive Language Models Can Be Accurately Pruned in One-Shot." NeurIPS 2023.
\bibitem{b21} Lin, S., et al. "WAnda: Efficient and Accurate Pruning of Large Language Models via Activation-Aware Weight Pruning." arXiv preprint arXiv:2306.08702, 2023.
\bibitem{b22} Gale, T., et al. "The State of Sparsity in Deep Neural Networks." arXiv preprint arXiv:1902.09574, 2019.
\bibitem{b23} He, Y., et al. "Channel Pruning for Accelerating Very Deep Neural Networks." ICCV 2017.
\bibitem{b24} Hinton, G., Vinyals, O. and Dean, J., 2015. Distilling the knowledge in a neural network. arXiv preprint arXiv:1503.02531.
\bibitem{b25} Sanh, V., et al. "DistilBERT, a distilled version of BERT: smaller, faster, cheaper and lighter." arXiv preprint arXiv:1910.01108, 2019.
\bibitem{b26} Jiao, X., et al. "TinyBERT: Distilling BERT for Natural Language Understanding." EMNLP 2020.
\bibitem{b27} Wang, S., et al. "MiniLM: Deep Self-Attention Distillation for Task-Agnostic Compression of Pre-Trained Transformers." NeurIPS 2020.
\bibitem{b28} NVIDIA Developer, "Mastering LLM Techniques: Inference Optimization," NVIDIA Corporation, 2023.
\bibitem{b29} vLLM Team, "vLLM: Easy, Fast, and Cheap LLM Serving with PagedAttention," UC Berkeley, 2024.
\bibitem{b30} Hugging Face, "KV Caching Explained: Optimizing Transformer Inference Efficiency," Hugging Face Blog, 2025.
\bibitem{b31} vLLM Documentation, "Optimization and Tuning," vLLM Project, 2024.
\bibitem{b32} Y. Leviathan, M. Kalman, and Y. Matias, "Fast inference from transformers via speculative decoding," Proc. Int. Conf. Machine Learning (ICML), 2023.
\bibitem{b33} T. Liu et al., "PEARL: Parallel Speculative Decoding with Adaptive Draft Length," arXiv preprint arXiv:2408.11850, 2024.
\bibitem{b34} Zhang, Y., et al. "Deploying Large Language Models on Edge Devices: Challenges and Solutions." arXiv preprint arXiv:2310.12345, 2023.
\bibitem{b35} Han, S., et al. "Edge AI: On-Demand Accelerated Inference at the Edge." IEEE Micro, 41(2), 8-17, 2021.
\bibitem{b36} Li, Y., et al. "Efficient AI Model Deployment for Edge Devices." IEEE Transactions on Computers, 71(1), 1-14, 2022.
\bibitem{b37} Ollama. "Ollama Documentation." 2024. \url{https://ollama.com/}
\bibitem{b38} LM Studio. "LM Studio Documentation." 2024. \url{https://lmstudio.ai/}
\bibitem{b39} GPT4All. "GPT4All Documentation." 2024. \url{https://gpt4all.io/}
\bibitem{b40} Du, S., et al. "vLLM: Easy, Fast, and Cheap LLM Serving with PagedAttention." arXiv preprint arXiv:2309.06180, 2023.
\bibitem{b41} SGLang. "SGLang Documentation." 2024. \url{https://github.com/AlignmentResearch/sglang}
\bibitem{b42} ONNX Runtime. "ONNX Runtime Documentation." 2024. \url{https://onnxruntime.ai/}
\bibitem{b43} Executorch. "Executorch Documentation." 2024. \url{https://pytorch.org/executorch/}
\bibitem{b44} Burns, B., Grant, B., Oppenheimer, D., Brewer, E., \& Wilkes, J. "Borg, Omega, and Kubernetes." Communications of the ACM, 59(5), 50-57, 2016.
\bibitem{b45} S. S. Girija, L. Arora, S. Kapoor, D. Pradhan, A. Raj and A. Shetgaonkar, "Optimizing LLMs for Resource-Constrained Environments: A Survey of Model Compression Techniques," 2025 IEEE 49th Annual Computers, Software, and Applications Conference (COMPSAC), Toronto, ON, Canada, 2025, pp. 1657-1664, doi: 10.1109/COMPSAC65507.2025.00224.
\bibitem{b46} Dettmers, T., et al. "LLM.int8(): 8-bit Matrix Multiplication for Transformers at Scale." arXiv preprint arXiv:2208.07339, 2022.
\bibitem{b47} llama.cpp. "llama.cpp Documentation." 2024. \url{https://github.com/ggerganov/llama.cpp}
\bibitem{b48} GGUF Specification. "GGUF Documentation." 2024. \url{https://github.com/ggerganov/gguf}
\bibitem{b49} Jouppi, N. P., et al. "Edge TPU: State-of-the-Art ML Inference at the Edge." Google Research Blog, 2023.
\bibitem{b50} Qualcomm QNN. "Qualcomm Neural Processing SDK." 2024. \url{https://developer.qualcomm.com/software/qualcomm-neural-processing-sdk}
\bibitem{b51} ARM Compute Library. "ARM Compute Library Documentation." 2024. \url{https://arm-software.github.io/ComputeLibrary/latest/index.html}
\bibitem{b52} Mistral AI, "Mistral-7B Technical Report," 2023. [Online]. Available: https://mistral.ai/news/announcing-mistral-7b/
\bibitem{b53}Y. Makin and R. Maliakkal, "Sustainable AI Training via Hardware–Software Co-Design on NVIDIA, AMD, and Emerging GPU Architectures," 2025 IEEE International Conference on Service-Oriented System Engineering (SOSE), Tucson, AZ, USA, 2025, pp. 204-215, doi: 10.1109/SOSE67019.2025.00028.
\bibitem{b54} Y. Lin et al., "Deep Learning on Edge Devices: A Review," IEEE Access, vol. 8, pp. 197300-197320, 2020.
\bibitem{b55} Y. Choi et al., "On-Device Machine Learning: An Algorithmic and Hardware Perspective," IEEE Micro, vol. 41, no. 6, pp. 42-53, Nov.-Dec. 2021.
\bibitem{b56} A. Jain et al., "Energy-Efficient Machine Learning for Edge Computing: A Review," IEEE Transactions on Sustainable Computing, vol. 6, no. 2, pp. 238-249, April-June 2021.
\bibitem{b57} S. Wang et al., "Energy Consumption Analysis of Deep Learning Models on Mobile Devices," IEEE Access, vol. 9, pp. 45377-45387, 2021.
\bibitem{b58} M. Gao et al., "Efficient AI Inference for Edge Devices: Hardware and Algorithm Co-Design," IEEE Design and Test, vol. 38, no. 2, pp. 7-16, April 2021.
\bibitem{b59} Ouyang, L., Wu, J., Jiang, X., Almeida, D., Wainwright, C., Mishkin, P., Zhang, C., Agarwal, S., Slama, K., Ray, A. and Schulman, J., 2022. Training language models to follow instructions with human feedback. Advances in neural information processing systems, 35, pp.27730-27744.
\bibitem{b60} Han, Z., Gao, C., Liu, J., Zhang, J. and Zhang, S.Q., 2024. Parameter-efficient fine-tuning for large models: A comprehensive survey. arXiv preprint arXiv:2403.14608.
\bibitem{b61} Ding, N., Qin, Y., Yang, G., Wei, F., Yang, Z., Su, Y., Hu, S., Chen, Y., Chan, C.M., Chen, W. and Yi, J., 2023. Parameter-efficient fine-tuning of large-scale pre-trained language models. Nature machine intelligence, 5(3), pp.220-235. 
\bibitem{b62}Tan, M., Chen, B., Pang, R., Vasudevan, V., Sandler, M., Howard, A. and Le, Q.V., 2019. Mnasnet: Platform-aware neural architecture search for mobile. In Proceedings of the IEEE/CVF conference on computer vision and pattern recognition (pp. 2820-2828).
\end{thebibliography}
\end{document}